\documentclass[conference]{ieeeconf}
\IEEEoverridecommandlockouts

\usepackage{cite}\usepackage{amsmath,amssymb,amsfonts}
\usepackage{algorithmic}
\usepackage{graphicx}
\usepackage{textcomp}
\usepackage{bm}
\usepackage{booktabs}
\usepackage{pifont}
\usepackage{caption}
\usepackage{xcolor}
\usepackage{tabularx}
\usepackage{hyperref}
\usepackage{hyperref} 
\hypersetup{
  colorlinks   = true,    
  urlcolor     = blue,    
  linkcolor    = red,    
  citecolor    = blue      
}

\def\BibTeX{{\rm B\kern-.05em{\sc i\kern-.025em b}\kern-.08em
    T\kern-.1667em\lower.7ex\hbox{E}\kern-.125emX}}

\begin{document}
\title{\textcolor{red}{C}%
\textcolor{red!70!orange}{o}%
\textcolor{red!50!orange}{L}%
\textcolor{red!30!orange}{I}: A Reproducible Platform for \textcolor{red}{C}ontinuum R\textcolor{red!70!orange}{o}bot \textcolor{red!50!orange}{L}earning via Monolithic 3D Printing and \textcolor{red!30!orange}{I}somorphic Teleoperation

\author{Ziyuan Tang, Chenxi Xiao$^*$}
\thanks{This work was supported by the Natural Science Foundation of Shanghai (Grant No. 25ZR1402370), and partially by Shanghai Frontiers Science Center of Human-centered Artificial Intelligence (ShangHAI), MoE Key Laboratory of Intelligent Perception and Human-Machine Collaboration (KLIP-HuMaCo).}
\thanks{Open source resources of this paper are released at \url{https://tangrobot.github.io/CoLI-website/}.}
\thanks{All authors are with the School of Information Science and Technology at ShanghaiTech University, Shanghai, China ($^*$corresponding author, {\tt\footnotesize tangzy2022, xiaochx@shanghaitech.edu.cn}).}
}
\maketitle

\begin{abstract}
Continuum robots offer strong potential for manipulation tasks due to their high degrees of freedom, compliant structures, and operational safety. However, their adoption in both research and practical applications has been hindered by reproducibility issues arising from complex fabrication and assembly processes, challenging kinematic modeling, and a lack of intuitive control interfaces. To address these challenges, we present a novel open-source continuum robot design. The platform features a simplified fabrication pipeline enabled by multi-material 3D printing, allowing the arm to be fabricated as a monolithic compliant structure with minimal assembly. Control is achieved through an isomorphic teleoperation interface that establishes a direct actuator-level mapping, eliminating the need for explicit kinematic modeling and providing a singularity-free mapping. Building on this hardware design, the platform further supports imitation-learning-based autonomous control. The proposed system is evaluated through hardware characterization and a set of manipulation tasks. Experimental results demonstrate that the platform provides a reproducible, learning-ready continuum robot system, accelerating algorithmic development and systematic benchmarking for the continuum robotics community.
\end{abstract}

\section{Introduction}

Continuum robots feature continuously deformable bodies and high-dimensional actuation, enabling dexterous manipulation in unstructured and confined environments\cite{7314984}. Inspired by biological appendages such as elephant trunks and octopus tentacles, these robots can conform to surrounding geometries, providing inherent safety through structural compliance and enabling navigation in cluttered spaces with their elongated, flexible morphology\cite{Kolachalama2020ContinuumRF,zhang2022continuumsurvey}. These properties make continuum robots well suited for applications where both safety and dexterity are critical, including minimally invasive surgery\cite{simaan2009design}, human--robot interaction\cite{POLYGERINOS2015135}, and exploration in confined or hazardous environments\cite{DONG2017218,CHANG2025103018}.

Despite their potential, continuum robots remain challenging to deploy for practical manipulation tasks, particularly in learning-based settings where repeated data collection is paramount. A primary limiting factor is the limited availability of accessible continuum robot hardware. Commercial continuum robots are primarily found in high-end surgical systems \cite{BerthetRayne2018i2Snake,Trevor2021tube}. Although alternative open-source solutions exist, many platforms still rely on custom-fabricated components that are expensive to manufacture and assemble\cite{10039108, 10145475, DONG2017218}, leading to high costs, low reproducibility, and consequently, barriers to deployment in practical scenarios\cite{9273083}. As a result, continuum robots are rarely used for large-scale data collection and robot learning, thereby limiting their adoption in data-driven manipulation research.

\begin{figure}[t]
    \centering
    \includegraphics[width=1\linewidth]{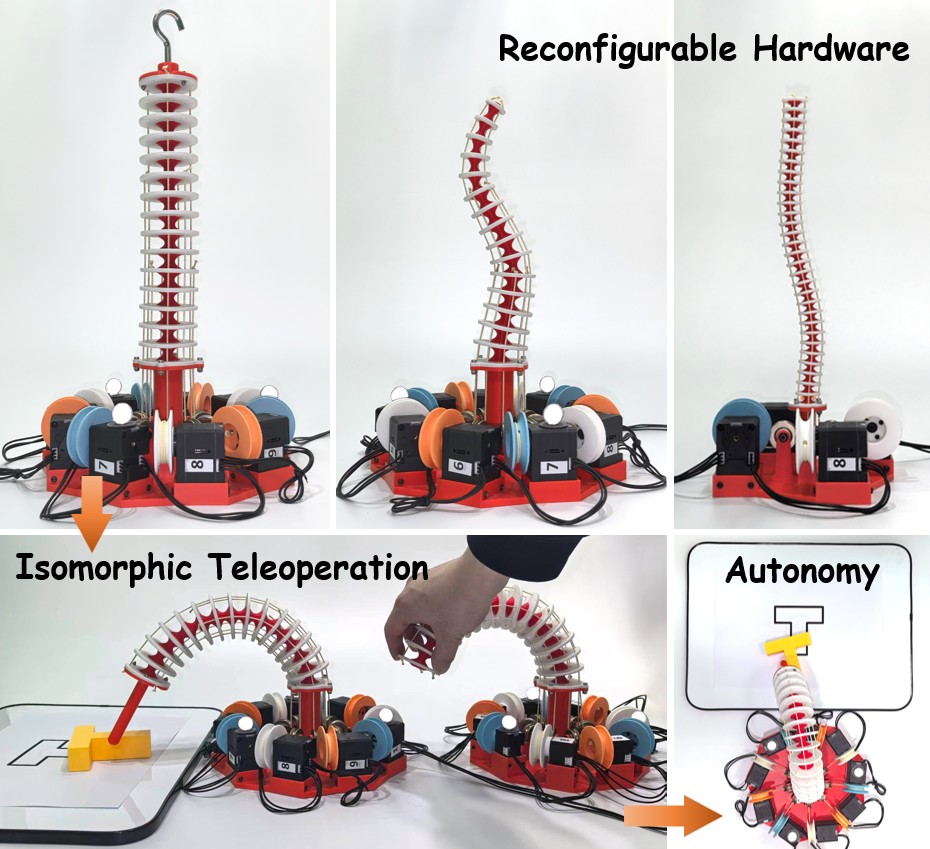}
    \caption{Overview of the proposed monolithic 3D-printed continuum robot platform with isomorphic teleoperation and imitation-learning-based autonomous control.}
    \label{fig:versions}
    \vspace{-4mm}
\end{figure}

Beyond hardware accessibility, controlling continuum robots introduces additional challenges. Their continuously deformable morphology and high-dimensional actuation result in significant actuation redundancy and strong coupling among degrees of freedom (DoFs) \cite{Kolachalama2020ContinuumRF}. Conventional end-effector-centric approaches typically rely on Cartesian control, regulating end-effector motion through kinematic models \cite{DONG2017218, 11247688}. However, such model-based formulations are difficult to derive accurately and calibrate, primarily due to the material compliance inherent in continuum robots \cite{George2018Strategies}.
Moreover, task-space control often leads to a mismatch between operator commands and the robot’s actuation space. The target pose specified by the operator may be unreachable because of discrepancies between the operator and robot workspaces, or it may fail to fully constrain the redundant DoFs of the follower robot. Consequently, the inherent actuation redundancy cannot be effectively exploited to modulate robot shape during manipulation \cite{zhang2022continuumsurvey}.

To address these challenges, two recent technological advances are particularly relevant. On the hardware side, the increasing availability of multi-material additive manufacturing has substantially lowered barriers to continuum robot fabrication\cite{Wehner2016SoftRobot}. Recent consumer-grade 3D printers can produce monolithic compliant structures with heterogeneous material properties and complex geometries in a single printing process\cite{BambuLabH2D}. This capability enables continuum robot designs that reduce assembly complexity, improve reproducibility, and support rapid design iteration, making them accessible to the broader community.

On the control side, isomorphic teleoperation has been widely adopted in the embodied AI community as an effective means of harnessing actuation redundancy\cite{wu2023gello,ben2024homie}. Such interfaces establish a one-to-one joint-wise correspondence between operator inputs and robot actuators, preserving the robot’s intrinsic actuation structure. By eliminating redundancy at the command level, isomorphic teleoperation produces actuator-aligned demonstrations suitable for motion replication and imitation learning\cite{An2025DexterousMT}. While this paradigm has proven effective for rigid-link robotic systems \cite{robotics9040102}, its application to continuum robots remains largely unexplored.

In this work, we leverage these advances and propose a continuum robot system that lowers both hardware and control barriers to learning-based manipulation. The robot is fabricated as a monolithic, multi-material 3D-printed structure that requires only tendon routing and actuator assembly during setup, which significantly reduces fabrication complexity and improves reproducibility. To fully exploit the robot’s high-dimensional actuation for control and data collection, we introduce an isomorphic teleoperation interface that enables intuitive full-body control while producing actuator-level demonstrations suitable for imitation learning. The system is integrated into the LeRobot framework \cite{cadene2024lerobot}, providing standardized pipelines for data collection, benchmarking, and learning-based control.
Building on this platform, we conducted systematic characterization and a set of benchmark tasks, including object manipulation and environment exploration under both teleoperated and autonomous conditions. The results demonstrate that the system provides a reproducible, learning-ready continuum robot platform, enabling data-driven research and algorithmic development in continuum robotics.

The main contributions of this work are:
\begin{enumerate}
    \item A monolithic multi-material fabrication paradigm that enables single-run, reproducible manufacturing of continuum robots for scalable research deployment.
    \item An isomorphic teleoperation framework that facilitates actuator-level teleoperation and demonstrations collection for imitation learning.
    \item A learning-ready continuum robot platform integrated with LeRobot framework, bridging continuum robot hardware and embodied AI pipelines.
    \item System-level validation across structure functionality, teleoperation, and autonomous manipulation, establishing a reproducible baseline for continuum robot learning.
\end{enumerate}

\section{Related Work}

\subsection{Fabrication of Continuum Robot}
The accessibility of a continuum robot depends on its fabrication approach, particularly given the structural requirements of its flexible backbone and rigid supporting components \cite{drones8060269}. The flexible backbone is commonly constructed using universal joints \cite{8665186}, springs \cite{8722729}, elastic metal rods \cite{simaan2009design}, and silicone materials \cite{10039108}, among others. Such designs typically require part-level fabrication, specialized material processing, and multi-step assembly procedures.

With the advancement of additive manufacturing technologies, more recent approaches have explored fabricating continuum robot components using 3D printing \cite{10522016,9273083,Li2025CHI}, improving accessibility and enabling customizable designs that were previously unattainable. However, most existing continuum robot designs still rely on producing rigid and flexible components separately, which requires subsequent assembly and alignment, introducing additional effort and potential dimensional errors. In contrast, our work aims to fabricate a dual-material monolithic structure, thereby further reducing assembly complexity and minimizing inaccuracies associated with manual fabrication.

\subsection{Teleoperation of Continuum Robots}
\label{related:teleoperation}

Teleoperation is a widely used approach for robot control. Traditional methods predominantly rely on task-space control, where operator inputs are mapped to task-space targets that are subsequently converted into actuator commands via kinematic models \cite{iyer2024open, Wang2024ARTeleoperation}. This task-space control paradigm has also been applied to continuum robots \cite{6385990,7487600}; however, it faces two primary challenges: workspace mismatch between the control interface and the robot, and the requirement for accurate deformable kinematic modeling.

Another line of work explores joint-space teleoperation \cite{WANG2024105804,6290272}, which enables direct control of individual actuators. This approach can exploit actuation redundancy and avoids the need for inverse kinematics. However, direct manual control of actuators is non-intuitive, and its usability is limited by the operator’s cognitive load.

Isomorphic teleoperation has emerged as a promising strategy to address these issues. By establishing a direct correspondence between operator inputs and robot actuators, it provides matched workspaces and improved control intuitiveness \cite{An2025DexterousMT}. Although this design has been widely adopted in recent embodied AI research for rigid-body robots \cite{wu2023gello,ben2024homie}, its application to continuum robots remains underexplored.

\begin{figure*}[t]
    \centering
    \includegraphics[width=1\linewidth]{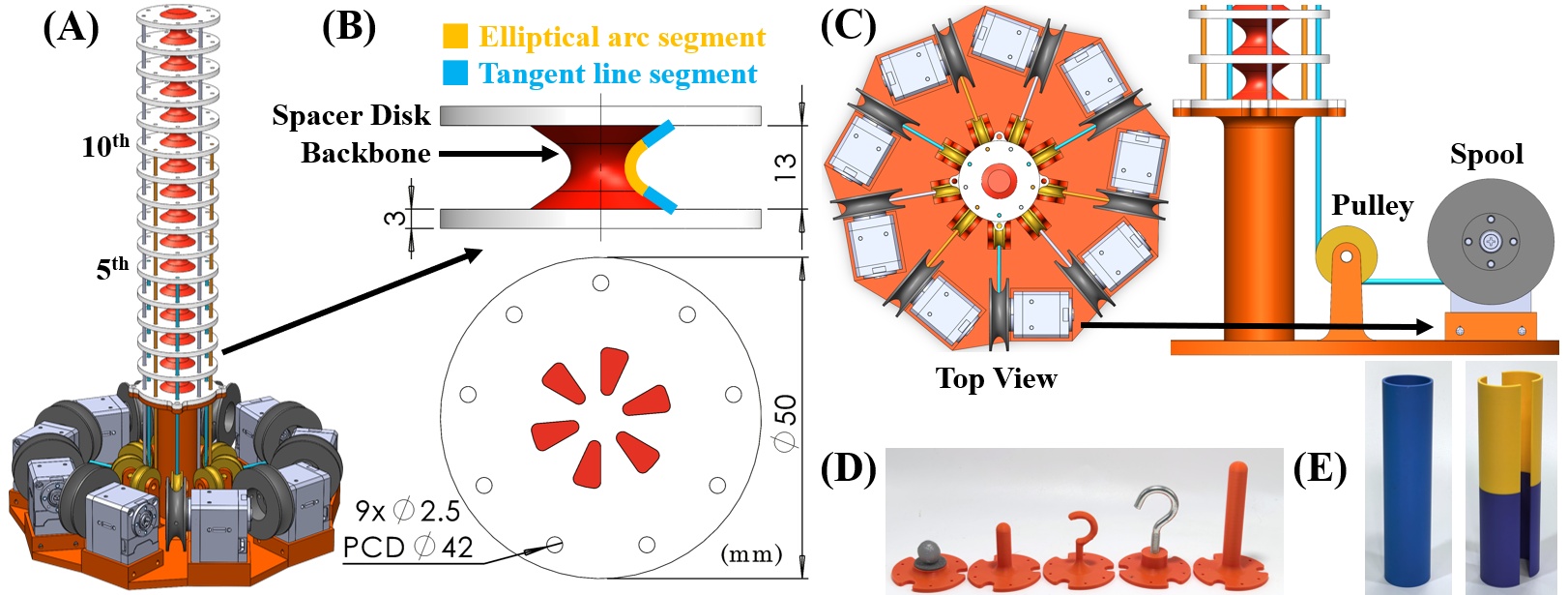}
    \caption{\textbf{Structure of the proposed continuum robot: }(A) structural body and cable routing, (B) joint structure between adjacent disks, (C) actuation mechanism, (D) end-effectors, and (E) calibration tools. }
    \label{fig:structure}
    \vspace{-3mm}
\end{figure*}

\subsection{Imitation Learning and Data-Driven Robot Manipulation}

Imitation learning has emerged as a powerful paradigm for robotic manipulation \cite{Fang2019Imitation}. By leveraging diffusion-based control \cite{chi2023diffusionpolicy} and transformer-based action models \cite{Zhao-RSS-23}, imitation learning has demonstrated strong performance across a wide range of manipulation tasks. A key challenge in enabling such autonomy is collecting high-quality demonstration datasets. Several open-source ecosystems have been developed to standardize data collection \cite{cadene2024lerobot,10611548} and benchmarking \cite{10611548,9001253}. Among them, the LeRobot framework provides unified pipelines for demonstration recording, imitation learning, and policy evaluation. However, existing frameworks predominantly target rigid-link manipulators, with limited support for continuum robots. This work aims to fill this gap by providing a continuum robot system specifically designed for scalable data collection and learning-based manipulation research.

\section{Methodology}
This section describes the proposed system design, including the continuum robot structure, fabrication process, teleoperation framework, and learning-based control pipeline.

\subsection{Monolithic 3D-Printable Continuum Robot}
To promote accessibility within the research community, we develop a novel continuum robot fabricated using a multi-material fused deposition modeling (FDM) 3D-printing process. The overall hardware design is illustrated in Fig.~\ref{fig:structure}~(A). The robot consists of a flexible backbone and 16 spacer disks, forming 15 serial joints that enable continuous bending. The robot has an overall height of 243~mm and a diameter of 50~mm. Details of the robot structure are described below.

\subsubsection{Flexible Backbone}
The flexible backbone is 3D printed from TPU~95A filament. It comprises a series of joints between disks, providing compliance and elastic restoring forces. Each joint is defined by planar elliptical arcs connected to tangent segments, which revolved around the central axis to form the three-dimensional geometry (Fig.~\ref{fig:structure}~(B)). The elliptical arcs create the thinnest regions, enabling compliance while producing restoring force during bending. Tangent segments constrain overhang angles within FDM printing limits and provide sufficient surface area for robust connections to the spacer disks.

\subsubsection{Spacer Disks}
\label{sec:curvature}

Spacer disks are printed together with the backbone using rigid PLA filament. Each disk has a diameter of 50~mm, a thickness of 3~mm, and nine evenly spaced holes for tendon routing. The center features a petal-shaped profile that mechanically interlocks with the flexible backbone, ensuring reliable structural coupling. The combination of a 50~mm disk diameter and a 13~mm backbone segment height constrains the joint curvature to $[0, 40]$, corresponding to a maximum inter-disk angle of $30^\circ$ when adjacent disks come into contact. Additional mounting holes are incorporated into the first and last disks to enable arm fixation and end-effector installation (Fig.~\ref{fig:structure}~(D)).

\subsubsection{3D Printing Process}
\begin{figure}[b]
    \vspace{-3mm}
    \centering
    \includegraphics[width=1\linewidth]{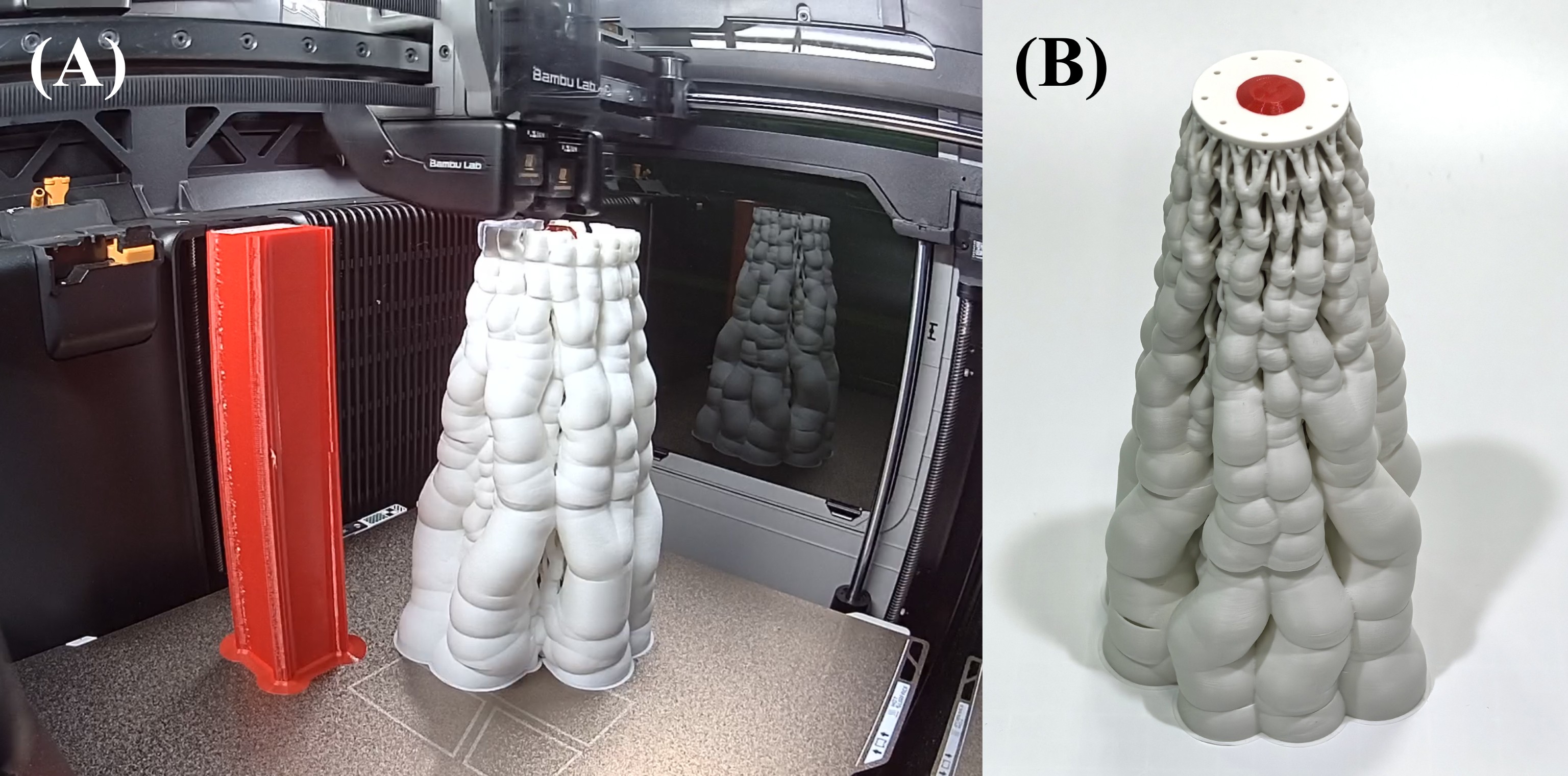}
    \caption{(A) The printing process of the continuum robot. (B) The printed continuum robot with support.}
    \label{fig:print_process}
\end{figure}

The proposed continuum robot is fabricated using a Bambu Lab H2D printer\cite{BambuLabH2D}, which supports monolithic multi-material printing with both rigid (PLA) and flexible (TPU) filaments. The printing process takes around 29~hours, with 0.2~mm layer height, 5 wall layers and 25\% infill for PLA components, and 100\% infill for TPU backbones. After printing, support material in overhanging regions can be removed within 15~minutes to obtain a fully functional continuum robot body. The printing process and fabricated prototype are shown in Fig.~\ref{fig:print_process}. The proposed design can be produced in a single printing run without human intervention, significantly simplifying the fabrication process compared to previous research \cite{drones8060269}.

Although this section describes a nominal design, the number of joints and spacer disks, as well as the overall height and diameter, can be readily customized to meet task-specific requirements. Benefiting from the flexibility and simplicity of 3D printing, the proposed fabrication process also enables rapid prototyping of diverse continuum robot morphologies. Two additional representative examples are shown in Fig.~\ref{fig:versions}: a slender variant suitable for enveloping grasps and a tower-like variant inspired by octopus tentacles.

\subsection{Actuation Mechanism}

Our continuum robot adopts a tendon-driven actuation architecture. All 15 joints are divided into three groups. Three tendons are assigned to each group and attached to the spacer disks located above the 5th, 10th, and terminal joints, respectively (Fig.~\ref{fig:structure}~(A)). Each tendon is routed downward through guide holes in the spacer disks, redirected by pulleys, and connected to a spool (Fig.~\ref{fig:structure}~(C)). The spools are driven by Dynamixel XC430-T240BB-T motors arranged circumferentially at the robot base. Notably, the number and arrangement of actuators are configurable to meet specific design requirements. For example, the slimmer variant shown in Fig.~\ref{fig:versions} employs only four actuators.

Our actuation strategy adopts independent tendon actuation. Each tendon is routed through the spacer disks and driven by an individual actuator, with one end anchored to the target disk and the other wound onto a motor-mounted spool. This configuration enables direct regulation of tendon tension via motor rotation and eliminates the need for manual tensioning during maintenance. Furthermore, it allows load sharing across multiple motors, thereby increasing payload capacity and improving overall system robustness.

\subsection{Isomorphic Teleoperation}
\label{sec:teleop}
To achieve concise and intuitive control, we propose an isomorphic teleoperation interface using a leader--follower scheme. The mechanical structure of the leader manipulator, including the continuum segment, tendon routing, and the placement of pulleys and spools, is identical to that of the follower robot. The key difference is that the leader uses Dynamixel XL330-M288-T motors with a lower reduction ratio, providing better backdrivability. This enables direct actuation by hand while maintaining sufficient pretension to preserve the robot’s shape without external support.

To ensure accurate motion synchronization, both the leader and follower robots are calibrated as follows:
\begin{enumerate}
\item Cylindrical sleeves (Fig.~\ref{fig:structure}~(E)) are tightly wrapped around the robots to maintain an upright configuration.
\item Motors are driven in PWM mode to apply tendon pretension. Absolute motor positions are recorded and defined as the zero reference for all actuators.
\end{enumerate}

\begin{figure}[t]
    \centering
    \includegraphics[width=1\linewidth]{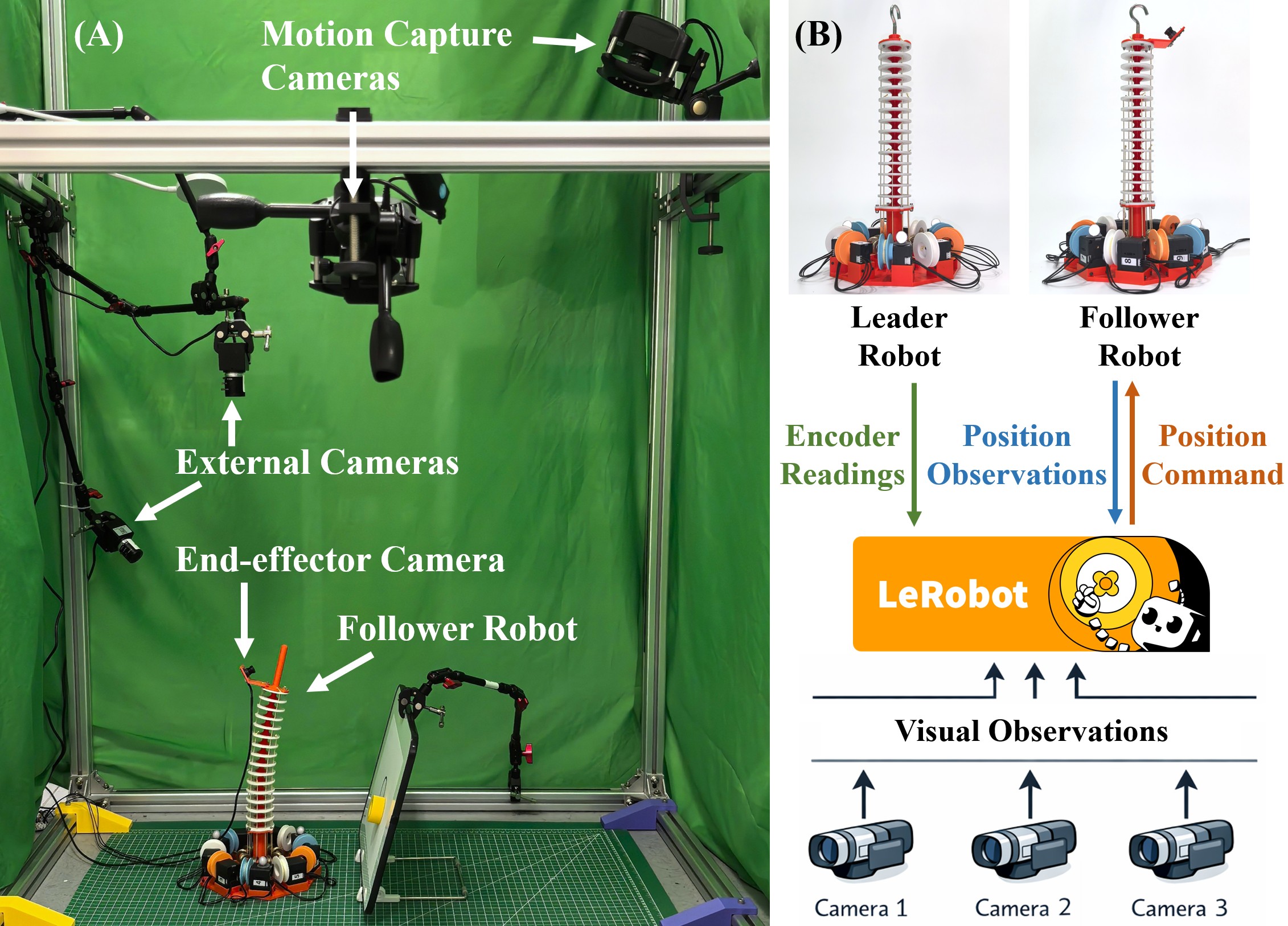}
    \caption{(A) Experimental setup for accuracy evaluation and imitation learning–driven manipulation tasks. (B) The data flow of learning–driven manipulation tasks.}
    \label{fig:environment}
    \vspace{-3mm}
\end{figure}

During calibration, the leader and follower robots operate at PWM mode with duty cycles of 25\% and 10\%, respectively, to achieve tendon pretension. These PWM values are empirically determined and may vary with different actuators or continuum robot designs.

During teleoperation, the leader motors operate in PWM mode (25\%) to maintain tendon tension, while the follower motors run in position control mode to track the leader’s joint-wise positions. When the user moves the leader, the resulting changes in tendon lengths (and thus motor encoder values) are mapped to target position commands for the follower motors, which rotate the follower spools accordingly to adjust tendon lengths. This mapping is implemented using a simple rule:
\begin{equation}
q^{\mathrm{f}}_i
=
\left( q^{\mathrm{l}}_i - q^{\mathrm{l}}_{i,0} \right)
+
q^{\mathrm{f}}_{i,0},
\end{equation}
where \(q^{\mathrm{l}}_i\) and \(q^{\mathrm{f}}_i\) denote the absolute encoder positions of the \(i\)-th motor on the leader and follower robots, respectively, and \(q^{\mathrm{l}}_{i,0}\) and \(q^{\mathrm{f}}_{i,0}\) are the corresponding zero offsets obtained during calibration. In each cycle, encoder readings of the servo motors on the leader robot are acquired synchronously, mapped to joint commands, and then synchronously written to the servo motors on the follower robot. This method does not require solving robot kinematics, allowing the system to operate with minimal delay without computational overhead.

\subsection{Imitation Learning-based Control}
Autonomous control has rarely been demonstrated on continuum robot platforms. To enable this capability, we leverage imitation learning from demonstrations collected via our aforementioned leader--follower interface (Sec.~\ref{sec:teleop}). To enhance task perception, the robot is additionally equipped with an end-effector camera and two external cameras, as shown in Fig.~\ref{fig:environment}. The entire system is integrated into the LeRobot framework~\cite{cadene2024lerobot}, which provides a unified pipeline for data collection, model training, and task execution, allowing future users to reproduce the same process easily.

During data collection, the system records both the target and current motor positions of the follower robot at each time step, along with camera images. Motor positions are expressed as offsets relative to calibrated zero points, and the sampling rate is set to 30~Hz to match the video stream. After dataset collection, we trained an Action Chunking with Transformers (ACT) model~\cite{Zhao-RSS-23} for each task to predict robot motion from proprioceptive and visual inputs. The policy uses a ResNet-18 backbone with a 4-layer transformer encoder (model dimension 512, 8 heads) and a variational objective (latent dimension 32). Training was performed for 100 epochs using AdamW (learning rate $10^{-5}$, batch size 16) on single NVIDIA A40 GPU. All policies were trained independently for each task. Evaluation results are presented and discussed in Sec.~\ref{sec:task_evl}.

\section{Experiments}
This section evaluates the performance of the proposed continuum robot system. Specifically, we assess (i) the mechanical and structural properties of the continuum robot design, (ii) the effectiveness of the proposed isomorphic teleoperation interface, and (iii) the feasibility of imitation learning–based manipulation through  task experiments.

\begin{figure}[t]
    \centering
    \includegraphics[width=1\linewidth]{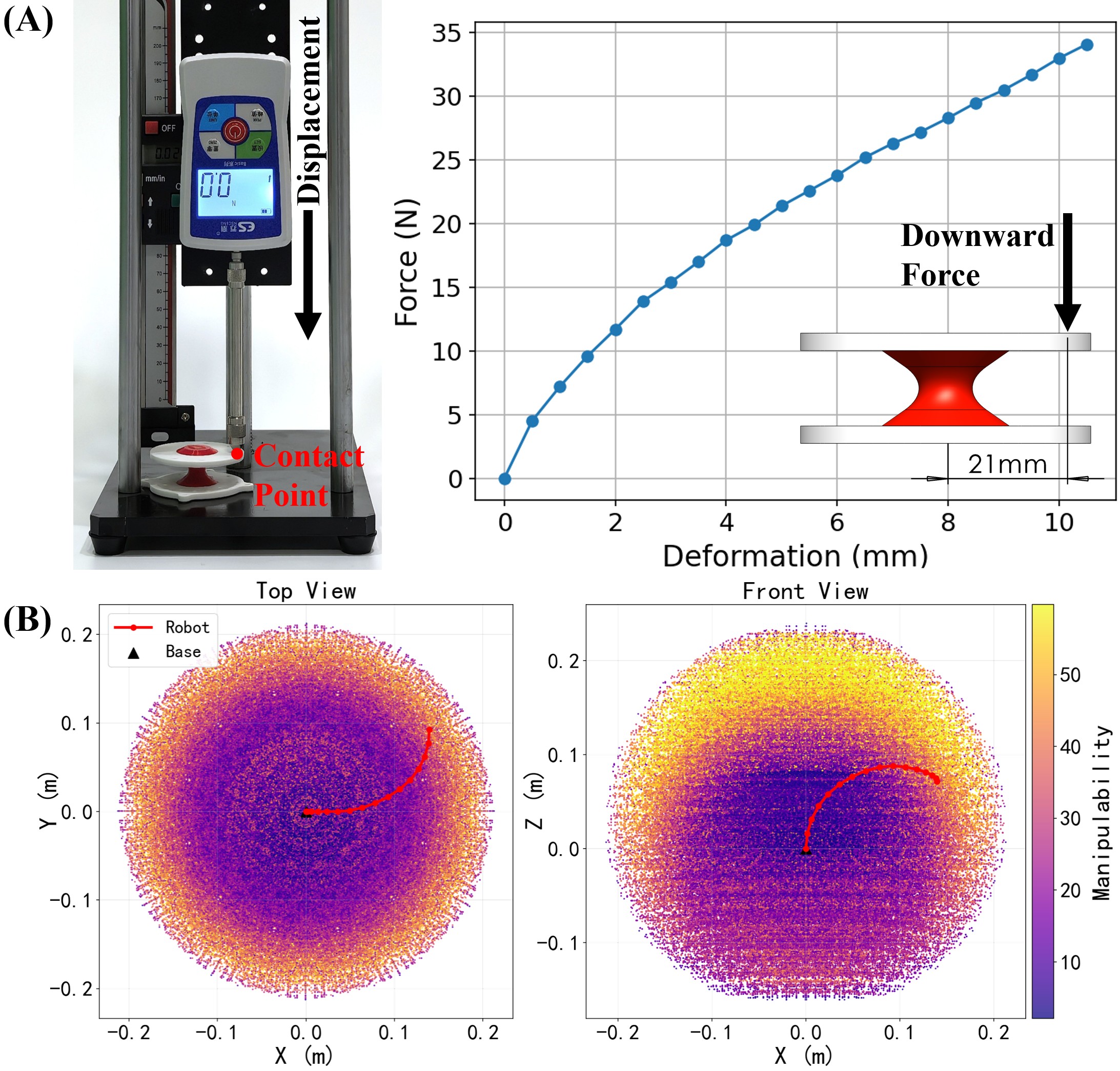}
    \caption{(A) Structural elasticity characterization, and (B) robot workspace analysis.}
    \label{fig:Experiment1}
    \vspace{-2mm}
\end{figure}

\subsection{Continuum Robot Performance Characterization}

\subsubsection{Force--Deformation Relationship}
We first characterize the force--deformation behavior of the monolithic 3D-printed continuum robot backbone. This is achieved by experimentally analyzing single-joint bending under external loading. The testing process involves a single-joint test sample mounted onto a test stand. The load was then applied by aligning the probe of the force gauge with the tendon-routing hole of the spacer disk, as illustrated in Fig.~\ref{fig:Experiment1}~(A).

As the probe moved downward, the joint underwent compressive deformation and exerted a restoring force on the probe, which was recorded by the force gauge. Measurements were sampled at 0.5~mm displacement increments until mechanical contact occurred between the upper and lower spacer disks, resulting in the force--deformation curve shown in Fig.~\ref{fig:Experiment1}~(A). The results indicate an approximately linear relationship between deformation and restoring force, with an approximate slope of 3.2~N/mm, and slightly larger restoring force generated in the small-deformation regime (less than 1~mm). This elasticity offers robot characteristics similar to other spring-based designs \cite{10081306}, for which existing modeling and control theories are applicable.

\subsubsection{Workspace}
We also evaluated the robot's workspace and manipulability, which are critical for task performance. For this purpose, the robot kinematic model was constructed using the Piecewise Constant Curvature (PCC) approach~\cite{Robert2010PCC}, with each joint curvature constrained to the range of $[0, 40]$ (Sec.~\ref{sec:curvature}). By randomly sampling bending directions and curvatures for each actuated joint, we computed the reachable workspace (shown in Fig.~\ref{fig:Experiment1}~(B)), which is approximately a
430~mm diameter sphere. Each sampled point is colored by the manipulability index, defined as
$w = \sqrt{\det(\mathbf{J}\mathbf{J}^T)}$, 
where $\mathbf{J} = \partial \mathbf{x} / \partial \boldsymbol{\ell}$ is the Jacobian of Cartesian end-effector displacements with respect to tendon length variations. This metric characterizes the local isotropy and effectiveness of actuation, with lower values indicating proximity to singular configurations and reduced controllability.

\subsubsection{Payload Capacity}
\begin{figure}[t]
    \centering
    \includegraphics[width=1\linewidth]{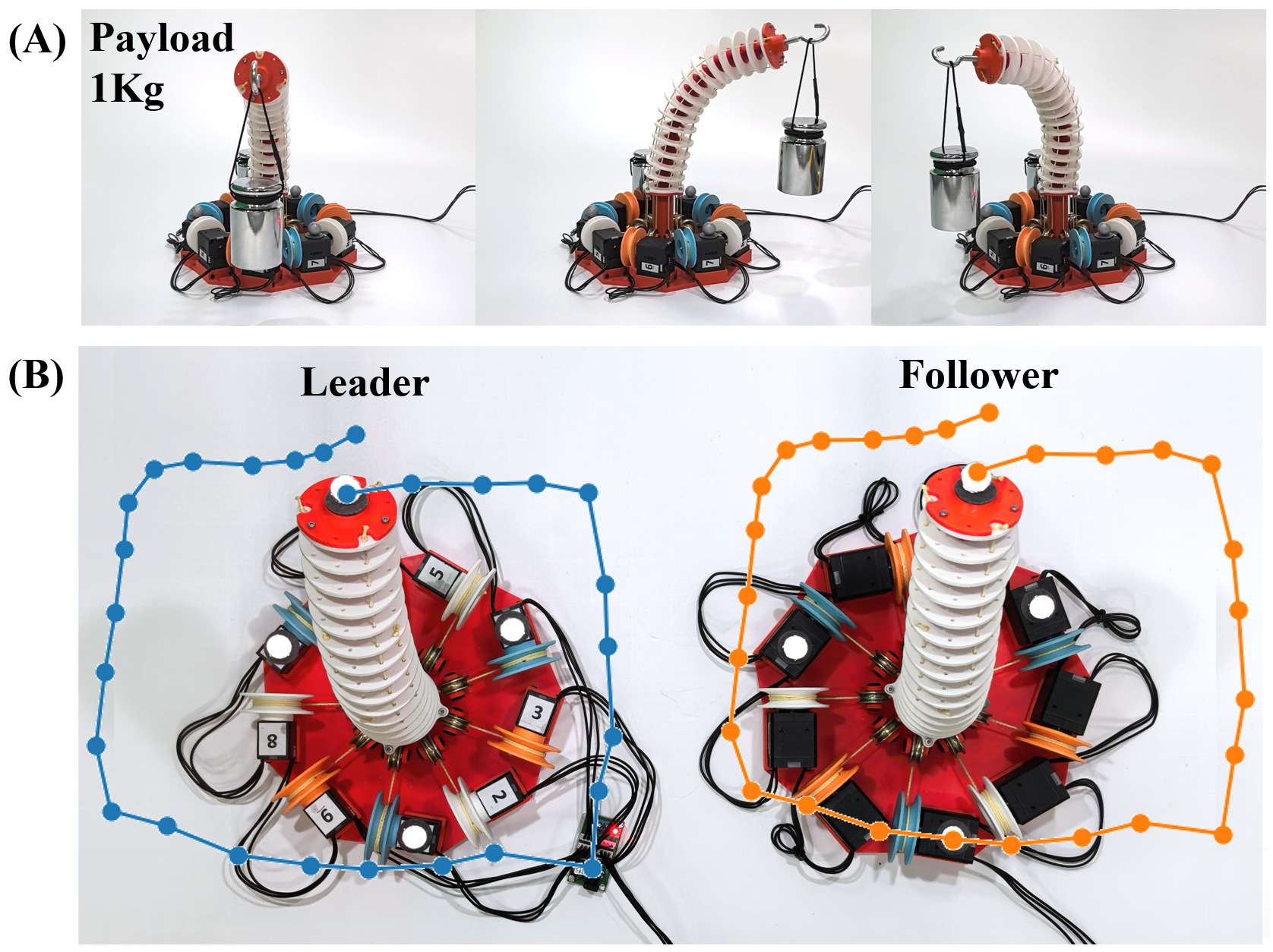}
    \caption{(A) Payload evaluation with a 1 kg load, and (B) teleoperation trajectory for both the leader and follower.}
    \label{fig:teleopration}
    \vspace{-2mm}
\end{figure}

To assess whether the continuum robot can support practical manipulation tasks, we evaluate its end-effector payload capacity through load-bearing experiments. A hook end effector was attached to the robot, and discrete weights were gradually suspended. The experiment showed that our robot can support a payload of up to 1\,kg. With this payload, the robot remained fully controllable under teleoperation conditions (i.e., capable of lifting the payload from the ground and executing diverse motions, as shown in Fig.~\ref{fig:teleopration}~(A)). When the payload reached 1.2\,kg, noticeable control degradation was observed, including increased tracking error and slower system response. Based on these observations, we define the maximum reliable end-effector payload of the proposed continuum robot as 1\,kg under dynamic operation.

\begin{figure*}[ht]
    \centering
    \includegraphics[width=1\linewidth]{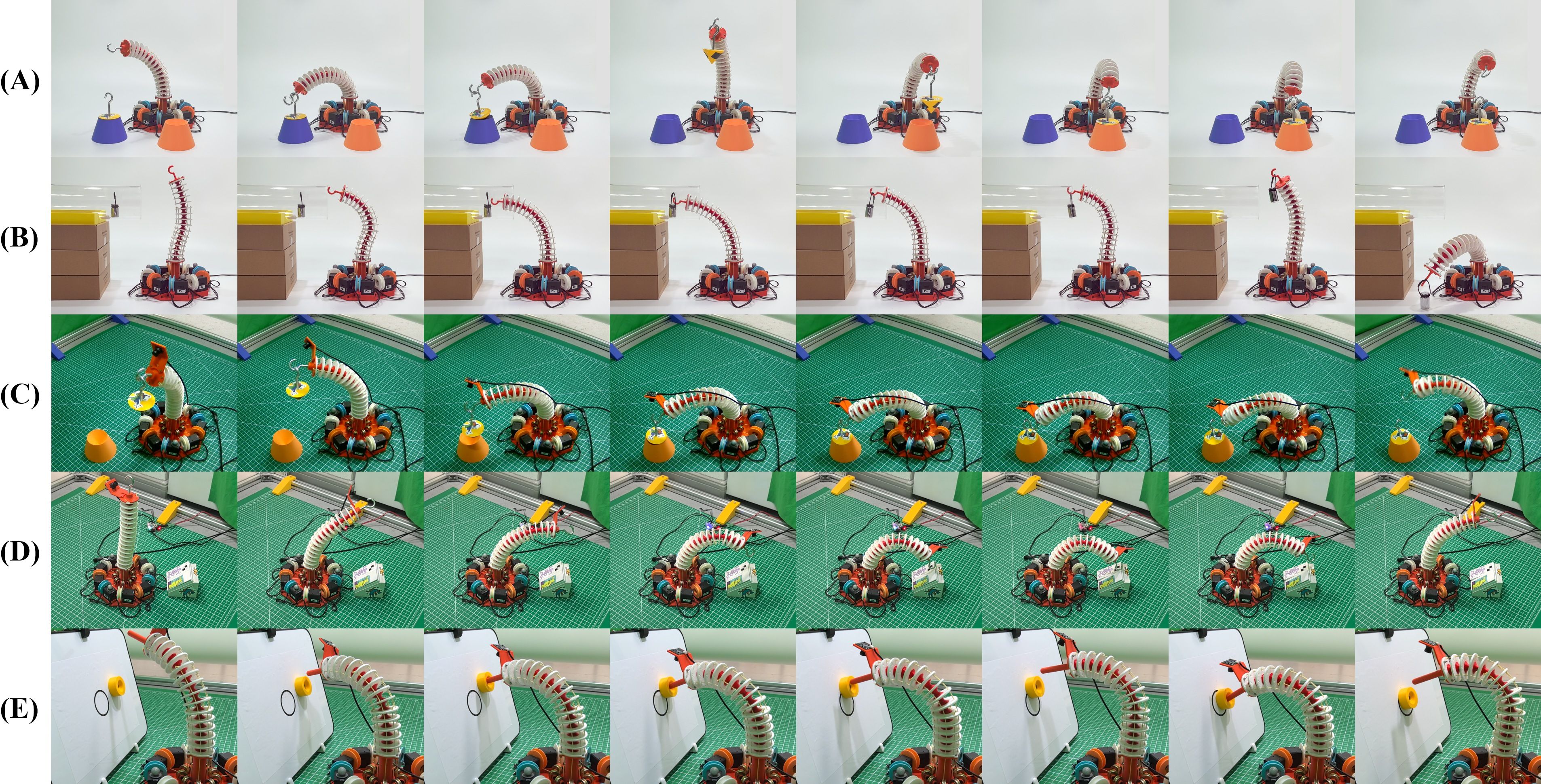}
    \caption{Demonstration of the robot’s \textbf{teleoperation dexterity} in (A) hooking and repositioning a weight and (B) reaching objects inside a tube. We further demonstrate \textbf{autonomous capabilities} in (C) positioning a weight, (D) contact-based switch toggling, and (E) non-prehensile pushing of an object toward a target location.}
    \label{fig:sequence}
    \vspace{-3mm}
\end{figure*}

\subsubsection{Durability}
In our experiments, the same pair of leader and follower robots was operated for over 15~hours in total. Throughout the experimental procedure, no structural damage, tendon slackening or degradation in control performance was observed. In addition, our robot can continuously operate for more than 90~minutes without failure. These results indicate that the robot's durability is sufficient for typical usage scenarios.

\subsection{Isomorphic Teleoperation Characterization}

\subsubsection{Tracking Accuracy}
\label{sec:teleop_position_acc}

Teleoperation requires the follower robot to accurately track the leader's position. We therefore first evaluate the teleoperation tracking accuracy of the proposed leader--follower framework. To estimate the relative pose between each end effector and its base, we used a motion capture system consisting of five Mars 1.3H cameras (Nokov Inc.) arranged around the workspace. Three reflective markers were mounted on the bases of both the leader and follower robots to define their respective base origins (as the centers of triangles), and one marker at the top of the robot to define the position of end effector. The relative positions of the leader and follower end-effector markers were manually aligned during mounting to minimize initial offsets.

We evaluated the isomorphic teleoperation by having the operator manually deform the leader robot to drive the follower, as shown in Fig.~\ref{fig:teleopration}~(B). Once both robots reached a target configuration and remained stationary, the spatial positions of their end effectors relative to their respective bases were simultaneously recorded. The tracking error was computed as the Euclidean norm of the difference.  Across 127 configurations, the mean positional tracking error was 8.2~mm (std: 3.3~mm), corresponding to 3.3\% of the 243~mm arm length. This accuracy was sufficient for all evaluated manipulation tasks.

\subsubsection{System Latency and Dynamic Response}
Using the same motion capture setup, we evaluated the dynamic performance of our teleoperation framework. Under a 30~Hz control frequency (synchronized with the camera frame rate), we recorded trajectories of both the user input and the robot output. We then characterized the dynamic performance by measuring the settling-time latency, defined as the time interval required for the follower to stabilize and align with the leader after the motion settles. Across 12 trajectories featuring diverse motion profiles, with peak end-effector velocities up to 5 cm/s, the average convergence latency was 158~ms. This settling time is relatively short and confirms that the follower can effectively synchronize with the leader’s intent, satisfying the requirements for high-fidelity demonstration collection.

\subsubsection{Evaluation of Robot Dexterity and Teleoperation Usability}

To assess both the dexterity of the continuum robot and the usability of the isomorphic teleoperation framework, we conducted two manipulation tasks: pick-and-place, and object retrieval in a confined space. The scene and motion sequences for both tasks are shown in Fig.~\ref{fig:sequence}~(A-B). Specifically, first, in the pick-and-place task shown in (A), the robot used a hook end effector to engage a target object on a circular pedestal and transfer it to another pedestal (object is also with a hook). Second, in the confined-space retrieval task shown in (B), the robot was required to enter a near-transparent cylindrical cavity with a 90~mm inner diameter, retrieve a 100~g hooked weight (positioned 8~cm from the cavity opening), and place it on the ground.

In both tasks, the operator successfully completed the manipulations via the leader--follower interface. The average completion times over five trials were 25.0~s for pick-and-place and 36.8~s for confined-space retrieval. These results confirm the usability of the teleoperation framework and highlight the continuum robot’s structural flexibility and capability for manipulation in constrained environments.

\subsection{Imitation Learning–Driven Manipulation}
\label{sec:task_evl}

Finally, we demonstrate the robot’s capability for imitation learning. Three autonomous manipulation tasks were performed using data collected via isomorphic teleoperation, followed by policy training with ACT \cite{Zhao-RSS-23} and real-robot evaluation. The success rates and average completion times of successful trials are summarized in Table~\ref{tab:task_performance}, with representative motion sequences shown in Fig.~\ref{fig:sequence}~(C-E).

\begin{table}[t]
\centering
\caption{Performance of imitation learning-driven manipulation tasks.}
\label{tab:task_performance}
\begin{tabular}{lcc} 
\toprule
Task & Success / Trials & Avg. Time (s) \\
\midrule
Object Placement          & 12 / 12 & 13.7 \\
Switch Toggling           & 10 / 12 & 15.1 \\
Magnetic Token Relocation & 10 / 12 & 22.0 \\
\bottomrule
\end{tabular}
\vspace{-2mm}
\end{table}

\subsubsection{Task 1: Object Placement}
The robot was equipped with a hook end-effector and initialized in a forward-bent configuration. At the start of each trial, a hooked target object was suspended from the end effector. The robot was required to autonomously localize a circular pedestal using vision, place the object onto the pedestal, disengage the hook, and return to an upright posture. Demonstrations were collected using both an end-effector camera and an overhead camera, yielding 80 trajectories with a total of 26,206 frames. During evaluation, each trial was terminated either after 60~s or upon task completion. The robot succeeded in all 12 evaluation trials. Notably, disengaging the hook is a delicate operation that poses a high risk of failure. In such challenging scenarios, our policy exhibited robust error recovery and retry behaviors until successful completion.

\subsubsection{Task 2: Switch Toggling}
The robot used a hook end-effector to press the left switch of a dual-switch panel and then return to an upright pose. Using the end-effector and overhead cameras, 61 demonstration trajectories with 20,423 frames were collected. Each evaluation trial was limited to 45~s. The robot succeeded in 10 out of 12 trials; the two failures were caused by incorrect press location.

\subsubsection{Task 3: Planar Pushing of a Circular Object}
The robot employed a rod-shaped end effector to push a circular object in the plane toward a target location. The object had a small central cavity, and was attached to a whiteboard tilted at $75^\circ$ relative to the ground via an internal magnet (to exploit the robot's available workspace). Task success was defined by the positional error of the object center relative to the target region center, with success defined to be within 30 mm threshold and task completion within 90~s. Demonstrations were collected using end-effector, overhead, and side-view cameras, yielding 69 trajectories and 39,586 frames. The robot succeeded in 10 out of 12 trials, with failures that the end-effector failed to engage object's central cavity.

Across all three tasks above, the robot consistently achieved success rates of at least 83\%, demonstrating the feasibility of imitation learning for continuum robots.

\subsection{Discussion}
With the goal of promoting continuum robots within the robot learning community, our system provides a reconfigurable and reproducible platform for continuum robot research. Overall, we believe that the proposed platform significantly lowers the barrier for researchers to conduct work on continuum robot systems. Nevertheless, several limitations remain. First, the maximum length of the monolithic continuum structure is constrained by the build volume of the 3D printer, which may restrict the arm's achievable workspace. Future work will explore modular or extendable designs, as well as optimized printing configurations, to improve scalability. Second, the primary cost of the system arises from the use of servo motors. To maintain sufficient payload capacity, we employ XC430-T240BB-T motors (USD~120 each), resulting in a total system cost of approximately USD~1{,}350 for both the leader and follower. Investigating lower-cost actuation alternatives could further improve accessibility, albeit with potential trade-offs in payload capability. Finally, we plan to incorporate richer sensing modalities and more diverse learning strategies, such as vision--language--action models, to evaluate system performance and usability in more complex learning-based manipulation scenarios.

\section{Conclusion}

This paper introduced a reconfigurable, reproducible, learning-ready continuum robot platform that unifies monolithic multi-material 3D printing, isomorphic teleoperation, and imitation-learning-based autonomy. 
The proposed design can be fabricated as a single compliant structure with minimal assembly, lowering the barrier to replicating continuum robot hardware and enabling rapid customization. Furthermore, our isomorphic teleoperation framework enables intuitive control without relying on kinematic modeling.
Experiments were conducted to characterize the platform’s performance, including force--deformation characteristics ($\sim$3.2~N/mm), reachable workspace (approximately a 430~mm diameter sphere), and dynamic payload capability (up to 1~kg). Building on the robot and the teleoperation interface, we enabled demonstration collection and deployed imitation-learning policies for multiple manipulation tasks, achieving robust autonomous behavior and failure-recovery motions. By establishing a standardized hardware and control baseline for continuum robot learning, our platform aims to accelerate reproducible research and community-driven progress in continuum robot learning.
Future work will focus on scalable hardware variants for longer structures, lower-cost actuation, and richer proprioceptive/tactile sensing, as well as broader evaluation in more diverse learning settings.

\section*{Acknowledgment}
The authors thank Kefei Wu and Zhe Xu for their assistance on ACT model training.

\bibliography{egbib}
\bibliographystyle{IEEEtran}
\end{document}